\documentclass{article}

    \PassOptionsToPackage{numbers, compress}{natbib}

\usepackage[preprint]{neurips_2023}

\usepackage[utf8]{inputenc} 
\usepackage[T1]{fontenc}    
\usepackage{hyperref}       
\usepackage{url}            
\usepackage{booktabs}       
\usepackage{amsfonts}       
\usepackage{nicefrac}       
\usepackage{microtype}      
\usepackage{xcolor}         
\usepackage{setspace}
\usepackage{graphicx}
\usepackage{enumitem}
\usepackage{multirow}
\usepackage{algpseudocode}
\usepackage[vlined,ruled,linesnumbered]{algorithm2e}
\usepackage{xcolor}
\usepackage{amsmath}
\usepackage{amssymb}
\usepackage{wrapfig}
\usepackage{tcolorbox}
\usepackage{makecell}
\tcbuselibrary{most}
\newtcolorbox{mybox}[2][]{colbacktitle=red!10!white, colback=gray!10!white,coltitle=black!70!black, title={#2},fonttitle=\bfseries,#1}

\title{HIPODE: Enhancing Offline Reinforcement Learning with High-Quality Synthetic Data \\from a Policy-Decoupled Approach}

\author{%
    Shixi Lian$^{1}$\thanks{Equal Contribution}~~, Yi Ma$^{1}$\footnotemark[1]~~, Jinyi Liu$^{1}$, \textbf{Yan Zheng}$^{1}$, \textbf{Zhaopeng Meng}$^{1}$ \\
    $^1$ College of Intelligence and Computing, Tianjin University, Tianjin, China \\
    \texttt{\{sidnee, mayi, jyliu, yanzheng, mengzp\}@tju.edu.cn}
}

\begin{document}

\maketitle

\begin{abstract}

Offline reinforcement learning (ORL) has gained attention as a means of training reinforcement learning models using pre-collected static data. To address the issue of limited data and improve downstream ORL performance, recent work has attempted to expand the dataset's coverage through data augmentation. However, most of these methods are tied to a specific policy (policy-dependent), where the generated data can only guarantee to support the current downstream ORL policy, limiting its usage scope on other downstream policies. Moreover, the quality of synthetic data is often not well-controlled, which limits the potential for further improving the downstream policy. To tackle these issues, we propose \textbf{HI}gh-quality \textbf{PO}licy-\textbf{DE}coupled~(HIPODE), a novel data augmentation method for ORL. On the one hand, HIPODE generates high-quality synthetic data by selecting states near the dataset distribution with potentially high value among candidate states using the negative sampling technique. On the other hand, HIPODE is policy-decoupled, thus can be used as a common plug-in method for any downstream ORL process. We conduct experiments on the widely studied TD3BC and CQL algorithms, and the results show that HIPODE outperforms the state-of-the-art policy-decoupled data augmentation method and most prevalent model-based ORL methods on D4RL benchmarks.
\end{abstract}

\section{Introduction}
\label{sec:intro}
Reinforcement learning (RL)   \cite{sutton2018reinforcement, precup2001off, sutton1991dyna} has achieved remarkable success in various applications, including game playing   \cite{mnih2013playing}, robotics   \cite{singh2022reinforcement}, finance   \cite{charpentier2021reinforcement}, and healthcare   \cite{liu2020reinforcement}. Nevertheless, traditional RL algorithms necessitate real-time interaction with the environment. In many real-world scenarios, such as autonomous driving   \cite{kiran2021deep} and medical treatment   \cite{liu2020reinforcement}, online learning is not feasible due to the high cost of failures, and collecting new data is often expensive or even dangerous   \cite{prudencio2023survey}. Consequently, Offline RL~(ORL)     \cite{lange2012batch} has garnered significant attention in recent years as it aims to learn from a dataset of previously collected experiences without further interaction with the environment.

In the offline setting, prior off-policy RL methods are known to fail on fixed offline datasets  \cite{haarnoja2018soft, fujimoto2018addressing}, even on expert demonstrations  \cite{fujimoto2019off}. The main reason of this could be the limited coverage of offline data. This can cause the policy visiting states that are out of the distribution~(OOD) of the dataset, and suffer from the extrapolation error on these states  \cite{fujimoto2019off,kumar2019stabilizing}. To alleviate extrapolation errors, most ORL researches attempt to avoid out-of-distribution states or actions, focusing on policy constraint   \cite{fujimoto2019off, wu2019behavior, liu2020provably,fujimoto_minimalist_2021}, support constraint   \cite{IQLKostrikovNL22, kumar2019stabilizing}, value regularization      \cite{kumar_conservative_nodate, ma2021conservative, ma2021offline, kumar2021dr3, kostrikov2021offline, an2021uncertainty}, and others. However, these approaches face the problem of the loss of generalization capability  \cite{lyu_double_2022}.

Different from mitigating the extrapolation error, data augmentation has been applied in ORL recently to expand the coverage of the dataset. The simplest approach is to add noise to the original dataset to obtain augmented data      \cite{sinha2022s4rl, weissenbacher2022koopman}, which could result in inaccurate dynamics transition that may not match the real environment. In contrast, dynamics models used in model-based RL can augment the dataset by rolling out synthetic samples. Inspired by this, existing works use the forward or backward dynamics models      \cite{yu2021combo, yu2020mopo, kidambi2020morel, lyu_double_2022, wang2021offline, wang2022bootstrapped, lu2021revisiting, guo2022model, rigter2022rambo, fucloser} to generate synthetic data and incorporate them into the policy training process. However, most of these methods are policy-dependent since they have to explicitly deal with unreliable data derived from inaccurate models to adapt to the downstream policy, thus limiting their data's application to augment other ORL algorithms.  Among them,    \cite{wang2021offline,lyu_double_2022} achieve policy-decoupled data augmentation. However, these methods lack explicit constraints to ensure the quality of the generated data, making the underlying mechanism by which they work unclear and limiting the potential of further improvement to the downstream policy.
 
To overcome the above-mentioned issues, we investigate the data augmentation method that is not dependent on the downstream ORL policy, which also ensures the quality of generated data. We first empirically analyze that high-quality data is beneficial for enhancing ORL performance. Then, we propose the \textbf{HI}gh-quality \textbf{PO}licy-\textbf{DE}coupled (HIPODE) data augmentation approach.


\begin{wrapfigure}{r}{0cm}
	\centering
        \includegraphics[width=0.525\textwidth]{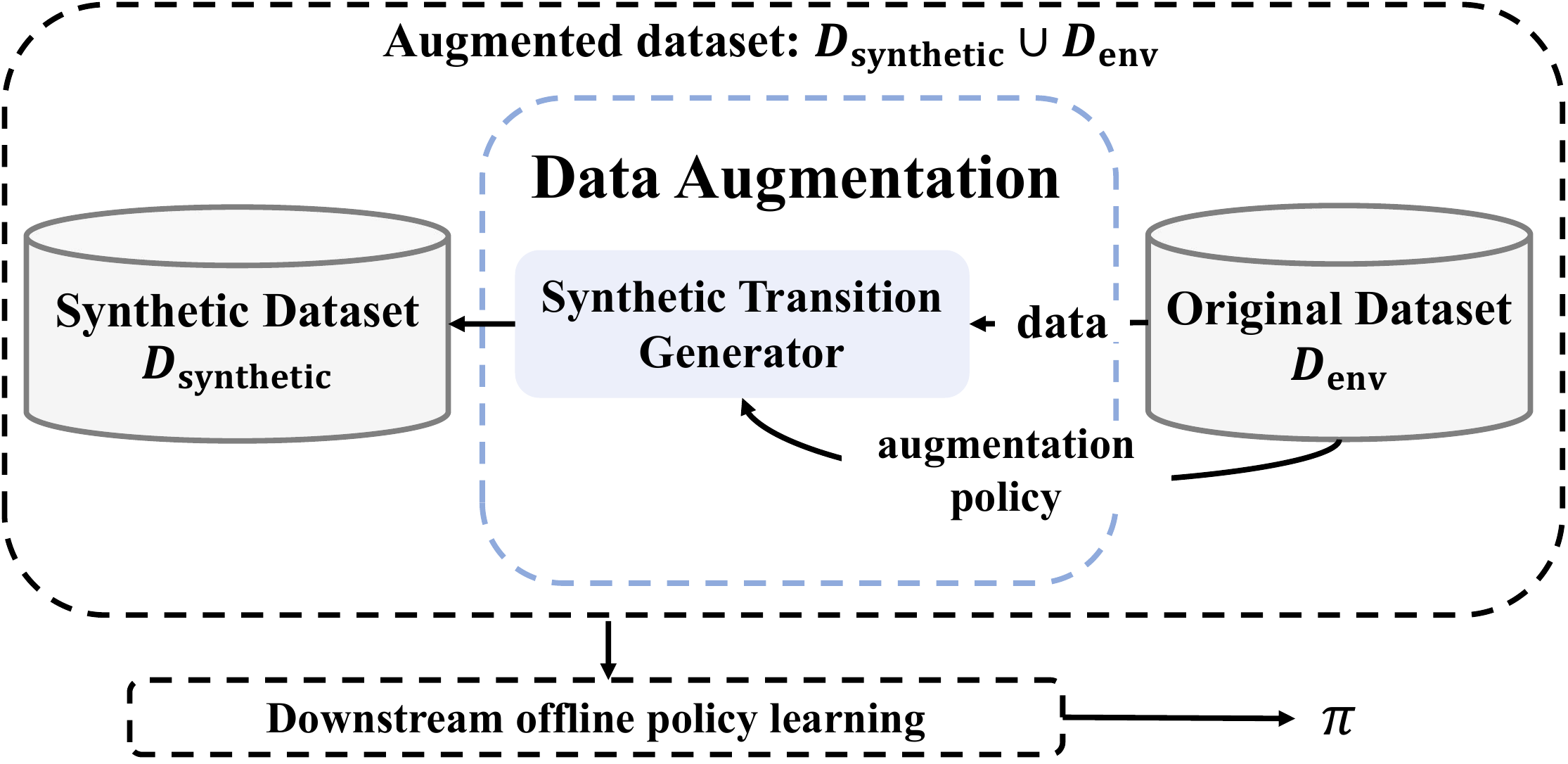}
	\caption{Outline of our data augmentation for ORL.}
	\label{fig:intro}
\end{wrapfigure}
We present the outline of our policy-decoupled data augmentation process for ORL in Fig.\ref{fig:intro}, which involves using specific augmentation policy to generate synthetic datasets based on the original dataset. These synthetic datasets are then used to expand the training data for any downstream ORL algorithm. Throughout this process, our key insight is to generate high-quality synthetic augmented data while ensuring authenticity (i.e. the proximity level between the synthetic data and the real data) as much as possible in such a policy-decoupled way. To this end, HIPODE uses a state transition model to generate next states and select the ones near the dataset distribution with potential highest value identified using a value network trained with negative sampling technique      \cite{luo2019learning}. Then the action and reward in the synthetic transition is replenished via inverse dynamics model and generative reward model, respectively. Note that data generation process of HIPODE does not need any information from the downstream policy, thus it's decoupled from the downstream offline policy learning process and can be used as a common plug-in method, similar to the role of data augmentation in Computer Vision (CV). Experimental results on D4RL benchmarks      \cite{fu2020d4rl} demonstrate that HIPODE significantly improves different baselines' performance and outperforms existing policy-decoupled data augmentation methods for ORL. To show the benefit of policy-decoupled approach, we also conduct experiments to show that synthetic data generated from HIPODE can benefit several downstream offline policy learning processes while existing policy-dependent augmentation methods may fail. 

To summarize, the contributions of this paper are:
\begin{itemize}[topsep=0pt, partopsep=0pt, leftmargin=*]
\item We investigate the impact of different types of augmented data on downstream ORL algorithms. Our findings indicate that high-quality data, as opposed to noisy data with high diversity, benefits downstream offline policy learning performance more.
\item We propose a novel policy-decoupled data augmentation method HIPODE for ORL. HIPODE serves as a common plugin that can augment high-quality synthetic data for any ORL algorithm, and is decoupled with downstream offline policy learning process.
\item We evaluate HIPODE on D4RL benchmarks and it significantly improves several widely used model-free ORL baselines. Furthermore, HIPODE outperforms state-of-the-art (SOTA) policy-decoupled data augmentation approaches for ORL.
\end{itemize}

\section{Related Work}

\textbf{Data augmentation in ORL.}  To address the challenge of limited data in ORL, various methods have been proposed to generate more sufficient data. Most of these approaches are policy-dependent, meaning they generate data based on the current policy and use it to refine the training of the same policy. These policy-dependent data augmentation methods can be divided into two categories. The first category seeks to generate pessimistic synthetic data that would be pessimistic enough if it is OOD, thus expand the dataset's coverage while mitigating the extrapolation error caused by such OOD data. The literature  \cite{yu2020mopo, kidambi2020morel} rely on the disagreement of dynamics ensembles or Q ensembles to construct a pessimistic MDP,  and  \cite{yu2021combo, rigter2022rambo, guo2022model} achieve the underestimation of synthetic data by unrolling the current policy in the model.
The second category does not explicitly pursue the underestimation of synthetic data. Among them,   \cite{fucloser} generates and selected synthetic data with low model disagreement, and BooT   \cite{wang2022bootstrapped} augments TT \cite{janner2021offline} with the synthetic data generated by itself. Besides, S4RL   \cite{sinha2022s4rl} and KFC~ \cite{weissenbacher2022koopman} add noise in a local area of states to smooth the critic. 

An obvious drawback of these aforementioned approaches, which are all policy-dependent, is that the generated data is closely related on the policy itself, causing that applying the generated data directly to the learning process of other policies is not guaranteed to perform well. To overcome this limitation, recent studies have explored policy-decoupled data augmentation techniques, which is the focus of this paper. Bi-directional rolling proposed in   \cite{lyu_double_2022} induce the double-check mechanism into offline data augmentation  ensure that the generated data is within the distribution and avoids inauthentic samples.  In \cite{wang2021offline}, a reverse dynamics model is proposed for ORL, and performs better than CQL   \cite{kumar_conservative_nodate} on maze environments. However, these two methods only consider the reliability of the synthetic data and neglect the quality of generated data, which may limit the performance. In contrast, HIPODE takes into account both the reliability and quality of the data. 

\textbf{Model-free ORL.}  Disregarding data augmentation, the model-free ORL algorithm investigates how to constrain the policy to approach the behavioral policy or support in static offline datasets. Existing methods implement this by policy constraint   \cite{fujimoto2019off, wu2019behavior, liu2020provably,fujimoto_minimalist_2021}, support constraint   \cite{IQLKostrikovNL22, kumar2019stabilizing}, value regularization      \cite{kumar_conservative_nodate, ma2021conservative, ma2021offline, kumar2021dr3, kostrikov2021offline, an2021uncertainty}, and others. Among them, we choose widely-used TD3BC   \cite{fujimoto_minimalist_2021} and CQL   \cite{kumar_conservative_nodate} to be the downstream policy learning algorithm to evaluate different data augmentation methods.

\section{What Kind of Augmented Data Can Help Improve ORL?}
\label{sec:motivation}

\begin{wraptable}{r}{8cm}
\vspace{-14px}
 \caption{Results of downstream ORL algorithm using different augmented data. Original denotes using the original dataset; Original + Diversity $\sigma$  and Original + Quality denote using high-diversity or high-quality augmented data; -r and -m-r denotes -random-v0 and -medium-replay-v0.
 }
 \vspace{6px}
\label{tab:augmenting_effects}
\centering
\scalebox{0.77}{
\begin{tabular}{l|lcc}
\toprule
\textbf{Task Name}                           & \textbf{Augmenting Type}     & \textbf{TD3BC}       & \textbf{CQL}                  \\ \midrule
\multirow{5}{*}{halfcheetah-r} & Original                  & 12.8                 & 17.0                 \\
                                    & Original + Diversity 0.01    & 12.1                 & 2.0                     \\
                                    & Original + Diversity 0.1     & 12.0                 & 16.1                     \\
                                    & Original + Diversity 1.0     & 9.2                  & 3.0                     \\
                                    & Original + Quality           &  25.8                & 23.8                     \\ \midrule
\multirow{5}{*}{halfcheetah-m-r}      & Original                  & 43.3                 & 42.5 \\
                                    & Original + Diversity 0.01 & 44.6 & 38.4 \\
                                    & Original + Diversity 0.1  & 44.3 &  1.8\\
                                    & Original + Diversity 1.0  & 41.8 &  25.6\\
                                    & Original + Quality        & 46.8 & 52.6 \\ \bottomrule
\end{tabular}
}
\end{wraptable}
As mentioned before, data augmentation methods in ORL often neglect the data quality. However, high-quality data are regarded as beneficial for learning~ \cite{fu2020d4rl}. Accordingly, we pose the question of whether the generation of high-quality data is also beneficial for ORL policies when compared to high-diversity data. We primarily investigate this question in this section.

To fairly investigate the effect of high-diversity data and high-quality data on the downstream ORL algorithm, we generate two types of data from real environments, instead of generating from other data augmentation techniques, to prevent potential bias due to inauthentic data impacting our findings. Moreover, to more closely match the offline setting, we limit the data generated in this section to only those that are not far away from the original dataset.
Concretely, we choose the following two types of augmentation policies: (1) \textbf{Policy of high diversity}, where random noise with different scales is added to the behavioral policy. Formally, $\pi_\text{noise}:=\mathcal{N}(a,\sigma \text{I}), \text{s.t.}, a\sim \pi_{\beta}$, where $\pi_\beta$ denotes the behavioural policy, $\mathcal{N}$ denotes the Gaussian distribution and $\text{I}$ denotes a identity matrix. The dataset after augmentation by this policy exhibits higher \textbf{diversity} compared to the original dataset. We refer to this type of method as `Diversity $\sigma$', where $\sigma$ belongs to {0.01, 0.1, 1.0}. (2) \textbf{Policy of high-quality}, a well-trained ORL policy, to ensure the action \textbf{quality}, i.e., return, derived from the ORL policy is similar to or higher than that of the actions in the dataset overall. Meanwhile, the generated data is also ensured to be close to the dataset. We refer this as `Quality'.

\begin{wrapfigure}{r}{0cm}
	\centering
 \vspace{-0.8cm} 
        \includegraphics[width=0.4\textwidth]{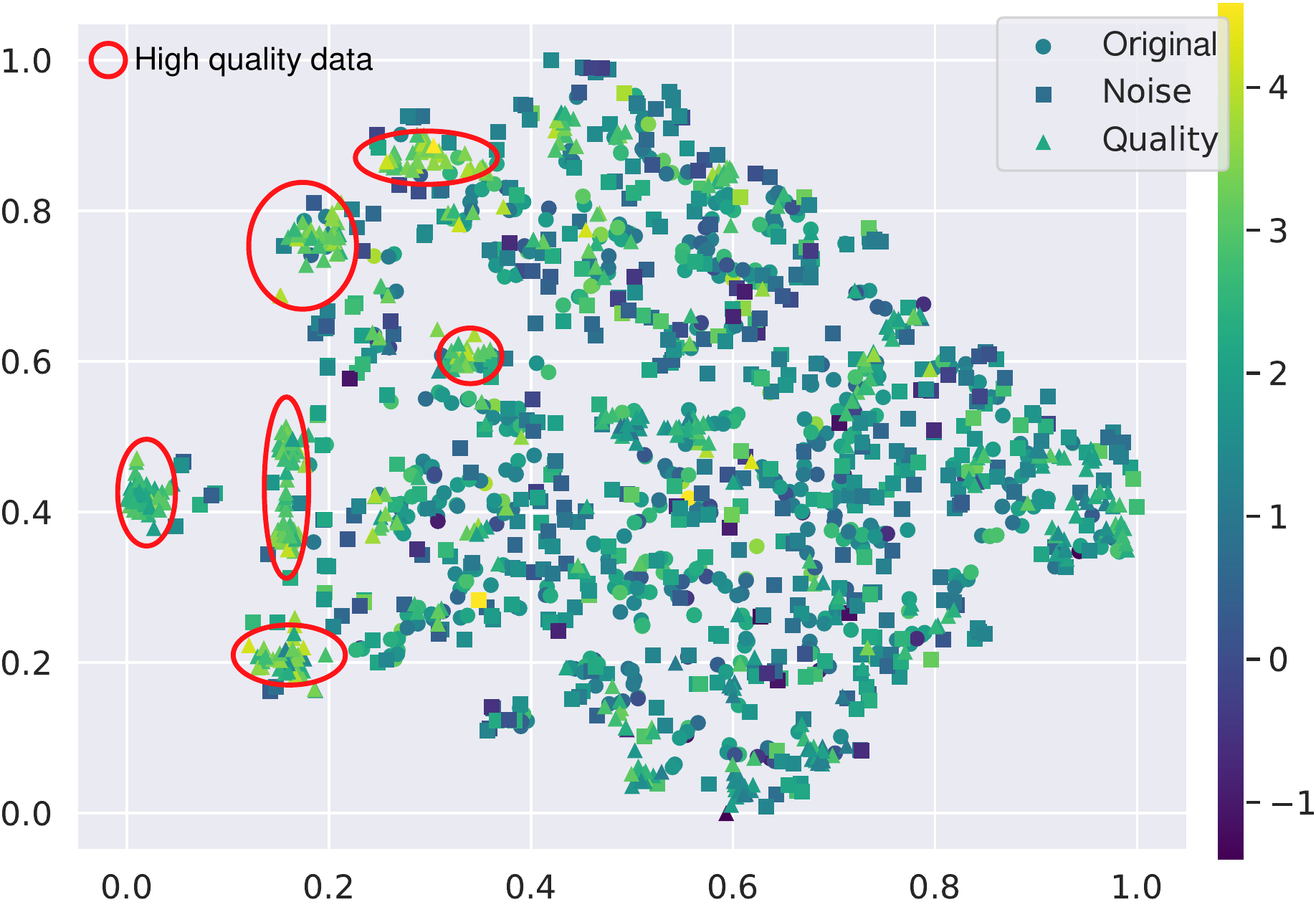}
	\caption{Action distributions of the original dataset, noise-policy-augmented data and quality-policy-augmented data. Brighter color indicates higher reward in a single time-step.}
	\label{fig:motivation}
\end{wrapfigure}
The augmented data and the original data are together used to train the downstream ORL algorithms. Normalized score reported in Table \ref{tab:augmenting_effects} shows that using ORL policy to augment data can always benefit down stream offline policy learning performance while using random noise policy may not. We further visualize the distribution of the original dataset, the noise-policy-augmented data, and the quality-policy-augmented data through t-Distributed Stochastic Neighbor Embedding (t-SNE)   \citep{hinton2002stochastic} in Fig.\ref{fig:motivation}. As we can see from it, compared with the distribution of the original data, the distribution of the noise-policy-augmented data is similar to the original dataset's while the distribution of the quality-policy-augmented data is relatively concentrated in several clusters. In addition, the quality-policy-augmented data indeed has higher rewards in a single time-step, as the color of the most triangle points are brighter. Based on these observations, we present the following takeaway: 

\begin{mybox}[detach title,before upper={\tcbtitle\quad}]{}
\textbf{Takeaway}: in the case that the augmented data is completely realistic, data with higher quality may be more beneficial than that of more diversity in improving downstream ORL algorithm.

\end{mybox}

\section{Method}
\label{sec:method}

\begin{figure}[tbp]
	\centering
        \includegraphics[width=1.0\textwidth]{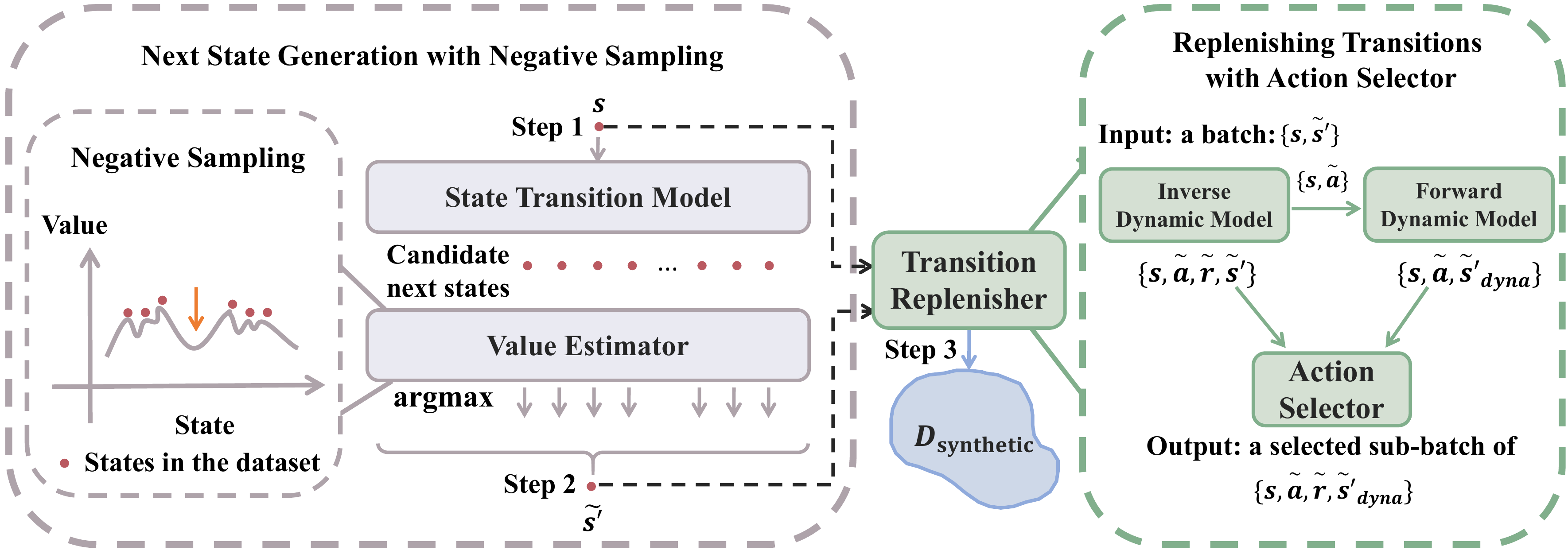}
	\caption{Illustration of HIPODE.}
	\label{fig:method}
\end{figure}

According to the takeaway above, we propose HIPODE to generate augmented data that maximizes its quality, i.e., return, while maintaining as much authenticity as possible in a policy-decoupled  way. We illustrate HIPODE in Fig.\ref{fig:method}. Specifically, given any state $s$, we first generate several candidate next states $\tilde{S}'_\text{cand}=\{\tilde{s}'_1, ..., \tilde{s}'_n\}$~(Step 1 in Fig.\ref{fig:method}). Then we select the one with the highest value as $\tilde{s}'$~(Step 2). 
Finally, given $s$ and $\tilde{s}'$, the action $\tilde {a}$ and the reward $\tilde {r}$ are produced using generative models~(Step 3), thus generating a transition $\{s,\tilde {a},\tilde {r},\tilde {s}'\}$. In the following, we introduce Step 1 and 2 in Section \ref{sec:state_generation} and Step 3 in Section \ref{sec:dc}. 

\subsection{Next State Generation with Negative Sampling}
\label{sec:state_generation}
Given a state $s$, we generate the next state through a state transition model, and filter the high-quality data for our purpose. In the following, we describe these two steps in detail.

\textbf{The forward state transition model.} We first train a state transition model $\tilde{p}_{\psi}(s'|s)$ to generate candidate next states. 
To guarantee the authenticity of generated next state, we model the state transition within the dataset with a conditional variational auto-encoder (CVAE) following  \cite{zhang2022state}, to ensure the generated next states are near the distribution of the dataset. Specifically, CVAE consists of an encoder and a decoder: the encoder takes the current state and the next state as input and manages to output an latent variable $z$ under the Gaussian distribution; the decoder takes $z$ and the current state as input and manages to map the latent variable $z$ to the desired space. We denote the encoder as $E_{\psi}(s,s')$ and the decoder as $D_{\psi}(s,z)$. The state transition model is then trained by maximizing its variational lower bound, which is equivalent to minimizing the following loss:
\begin{equation}\label{state transition loss}
    \mathcal{L}(\psi)=\mathbb{E}_{(s,s')\sim \mathcal{D}_\text{env}, z \sim E_{\psi}(s,s')}[(s'-D_{\psi}(s,z))^2+D_{\text{KL}}(E_{\psi}(s,s')\Vert\mathcal{N}(0,\text{I}))].
\end{equation}
where $\text{I}$ represents an identity matrix and $\mathcal{D}_{\text{env}}$ represents the original dataset. The first term of RHS of Eq.\eqref{state transition loss} represents the reconstruction loss where the approximated next state is decoded from $z$, given the current state. The second term of RHS represents the KL distance between the distribution of $z$ and the Gaussian distribution so that a sampled $z$ from a Gaussian distribution can be decoded to the desired state space when generating. Thus, given a state $s$, $n$ candidate next states $\tilde{S}'_\text{cand}=\{\tilde{s}'_1, ..., \tilde{s}'_n\}, \text{s.t., }\tilde{s}'_i\sim D_{\psi}(\tilde{s}'_i|s,z)$ are sampled.

\textbf{Value Approximation with Negative Sampling.} To filter out the generated next states and form synthetic transitions, a value approximator is trained using SARSA-style updating to predict the value of different states. Since the generated next states may not be present in the dataset, the negative sampling technique   \cite{Luo2020Learning} is employed to avoid overestimation of states outside the dataset. Specifically, for states within the dataset, standard TD-learning is performed as demonstrated in Eq.\ref{v td loss}:
\begin{equation}\label{v td loss}
    \mathcal{L}^\text{td}_{\theta}(s)=\mathbb{E}_{(s,r,s') \sim \mathcal{D}_\text{env}}[r+\gamma V_{\bar{\theta}}(s') - V_{\theta}(s) ]^2,
\end{equation}
where $V_{\bar{\theta}}$ is the target value function and $\gamma$ is the discount factor. Furthermore, to conservatively estimate the value of states outside the dataset distribution, we sample states around the dataset states by adding Gaussian noise and evaluate the L2 distances between the sampled noisy states and the original states. The greater the distance between the sampled state and the original state, the more severe the penalties imposed on the sampled state, as shown in Eq.\eqref{v ns loss}: 
\begin{equation}\label{v ns loss}
    \mathcal{L}^\text{ns}_{\theta}(s)=\mathbb{E}_{s\sim \mathcal{N}(s_\text{d},\sigma \text{I}), (s_\text{d},r,s') \sim \mathcal{D}_\text{env}}[r+\gamma V_{\bar{\theta}}(s')-\alpha \Vert s-s_\text{d} \Vert - V_{\theta}(s)]^2,
\end{equation}
where $s_\text{d}$ denotes the state sampled from the original dataset and $\alpha$ denotes the penalty weight. Thus, the optimization objective of the value approximator  to minimize is represented by Eq.\eqref{v loss}:
\begin{equation}\label{v loss}
    \mathcal{L}(\theta)=\mathcal{L}(\theta)^\text{td}+\mathcal{L}(\theta)^\text{ns}.
\end{equation}
After training, all the candidate next states $S'_\text{cand}$ are input into the value approximator to obtain their values. Then the candidate next state with the highest value estimation is selected, formally $\tilde{s}'=\text{argmax}_{\tilde{s}'_\text{cand}}V(\tilde{s}'_\text{cand}),\tilde{s}'_\text{cand}\in \tilde{S}'_\text{cand}$ . Intuitively, a state can be selected in two cases:
\begin{itemize}[topsep=0pt, partopsep=0pt, leftmargin=*]
   \item States within the dataset. This is because other candidate states  that not in the dataset are severely underestimated. In this case, the selected state can be considered reliable.
   \item States with high true value near the dataset distribution. Since its estimated value is significantly penalized during training, there is a high probability that a selected state close to the distribution has a high true value.
\end{itemize}
Therefore, by filtering the candidate next states generated by the state transition model using the value approximator, we can obtain the augmented next state with similar or higher quality than that in the datasets while maintaining as much authenticity as possible.

\subsection{Replenishing Transitions with Action Selector}\label{sec:dc}
Based on the selected high-quality next state, in this section, we aim to generate an authentic action that can lead the current state to the generated next state. Specifically, an inverse model $M_\text{inv}=\tilde{p}_{\epsilon}(a|s,s')$ is trained to generate actions conditioned on $s$ and the selected $\tilde{s}'$. Similar to the state transition model, we also use a CVAE for generating actions. We denote the encoder as $E_{\epsilon}(a,s,s')$ and the decoder as $D_{\epsilon}(s,s',z)$. The inverse model is then trained by maximizing its variational lower bound, which is equivalent to minimizing the following loss shown as Eq.\eqref{inverse dynamics loss}:
\begin{equation}\label{inverse dynamics loss}
    \mathcal{L}(\epsilon)=\mathbb{E}_{(a,s,s')\sim \mathcal{D}_\text{env}, z\sim E_{\epsilon}(a,s,s')}[(a-D_{\epsilon}(s,s',z))^2+D_{\text{KL}}(E_{\epsilon}(a,s,s')\Vert\mathcal{N}(0,\text{I}))].
\end{equation}
Besides, rewards are generated the same way as actions, using another model with encoder $E_\zeta(r,s,s')$ and decoder $D_\zeta(s,s',z)$.

Although the generated state have high quality and authenticity as described in Section \ref{sec:state_generation}, the action generated by the inverse dynamics model may be inauthentic, i.e. the generated action can not lead to the selected next state. Therefore, a filtering mechanism is imposed on actions for their reliability. We further draw on a forward dynamics model $M_\text{for\_dyna}=\tilde{p}_{w}(\tilde{s}'_\text{dyna}|s,a)$ representing the probability of the next state given the current state and action. The dynamics model is optimized by maximizing the log-likelihood of the static dataset, formally shown in Eq.\eqref{forward dynamics loss}: 
\begin{equation}\label{forward dynamics loss}
    \mathcal{L}(w)=\mathbb{E}_{(s,a,s')\sim \mathcal{D}_\text{env}}[-{\rm log}\ \tilde{p}_{w}(s'|s,a)].
\end{equation}
Combining the forward dynamics model $M_{\text{for\_dyna}}$ and the inverse dynamics model $M_{\text{inv}}$, an action is assumed to be reliable to lead to the selected state $\tilde{s}'$ when the distance between the selected state $\tilde{s}'$ and the forward-predicted state $\tilde{s}'_\text{dyna} = M_\text{for\_dyna}(s,\tilde{a})$ is small enough. In practice, instead of setting a threshold for measuring the distance between $\tilde{s}'$ and $\tilde{s}'_\text{dyna}$, we pick up in a batch $\lambda$-portion of the generated data with the lowest $\Vert \tilde{s}'_\text{dyna}-\tilde{s}'\Vert$ values and consider them the most reliable subset of the batch. Then this subset of data is used as the final augmented data. 

    
    

HIPODE is summarized in Algorithm \ref{main algorithm}, where $N(\mathcal{D}_\text{env})$ is the amount of data in the original dataset. Line 5-7 refers to selecting the next state with negative sampling, described in Section \ref{sec:state_generation} for generating synthetic next states that are close to the dataset distribution and have the potential for the highest value. Line 8-10 and 12 refer to replenishing transitions with an action selector, as described in Section \ref{sec:dc} for generating synthetic actions and rewards. Line 13-14 refers to merging the synthetic data with the original dataset for downstream offline policy training to obtain the offline policy $\pi$.
\begin{algorithm}[tbp]
\small
\caption{HIPODE}
\label{main algorithm}
    \textbf{Input:} Offline dataset $\mathcal{D}_\text{env}=\{(s,a,r,s')\}$, penalty wieght $\alpha$, synthetic rate $\eta$, action selecting rate $\lambda$, number of candidate next states $n$

    \textbf{Output:} Policy $\pi$

    Train state transition model $D_{\psi}(s,z)$ by minimizing Eq.\eqref{state transition loss}, value estimator $V_{\theta}(s)$ by minimizing Eq.\eqref{v loss}, dynamics model $\hat{p}_{w}(s'|s,a)$ by minimizing  Eq.\eqref{forward dynamics loss} and inverse action model $D_{\epsilon}(s,s',z)$ as well as inverse reward model $D_\zeta(s,s',z)$ by minimizing Eq.\eqref{inverse dynamics loss} and a similar loss for reward generation respectively
    
    \Repeat{reaching maximum generating amount, which is $\eta N(\mathcal{D}_\text{env})$ }{
        Sample a batch of $s$ from $\mathcal{D}_\text{env}$
        
        Sample $\tilde{S}'_\text{cand}$ containing $n$ $\tilde{s}'_i$ from $\tilde{s}'_i\sim D_{\psi}(s,z),z\sim \mathcal{N}(0,\text{I})$
        
       $\tilde{s}'=\text{argmax}_{\tilde{s}'_\text{cand}}V(\tilde{s}'_\text{cand}),\tilde{s}'_\text{cand}\in \tilde{S}'_\text{cand}$
        
        Sample actions $\tilde{a} \sim D_{\epsilon}(s,s',z),z\sim \mathcal{N}(0,\text{I})$ and sample rewards $\tilde{r} \sim D_{\zeta}(s,s',z),z\sim \mathcal{N}(0,\text{I})$
        
        Sample $\tilde{s}'_\text{dyna}$ from $\tilde{s}'_\text{dyna}\sim \tilde{p}_{w}(\tilde{s}'_\text{dyna}|s,a)$
        
    }
    Select top $\lambda$-portion authentic actions to construct $D_\text{synthetic}$ with least $\Vert \tilde{s}'_\text{dyna}-\tilde{s}'\Vert$
    
    Merge the synthetic dataset and the original dataset $D=D\cup D_\text{synthetic}$
    
    Use any model-free offline policy learning algorithm to obtain $\pi$

    \textbf{return} $\pi$
\end{algorithm}

\section{Experiments}
\label{sec:experiment}
In this section, we evaluate HIPODE based on two representative and widely-used offline policy learning algorithm: TD3BC   \cite{fujimoto_minimalist_2021} and CQL      \cite{kumar_conservative_nodate}.
We aim to answer these questions: 
\begin{itemize}[topsep=0pt, partopsep=0pt, leftmargin=*]
    \item[] \textbf{\textit{Q1}}: Can our proposed algorithm HIPODE improve existing ORL algorithms and exhibit consistent superiority in comparison to other data augmentation technique?  
    \item[] \textbf{\textit{Q2}}: Is augmenting synthetic data or high-quality synthetic data critical for ORL policy?
    \item[] \textbf{\textit{Q3}}: Does our policy-decoupled data augmentation algorithm HIPODE outperform the conventional policy-dependent data augmentation methods?
    \item[] \textbf{\textit{Q4}}: In HIPODE, what roles do the negative sampling and transition selector components play? 
\end{itemize}
%
In the following, we answer \textit{Q1} in Section \ref{main experiments}, showing the effectiveness and superiority of HIPODE by combining it with offline RL algorithms on MuJoCo     \cite{todorov2012mujoco} tasks. Then, we present an ablation study in details in Section \ref{ablation study} to answer \textit{Q2}. We answer \textit{Q3} in Section \ref{policy-dependent} by comparing HIPODE with policy-dependent data augmentation methods. Finally, we answer \textit{Q4} in Appendix \ref{appendix: additional results}.

\subsection{Performance on MuJoCo}\label{main experiments}

\textbf{Evaluation settings}. We demonstrate the benefits of HIPODE on D4RL MuJoCo-v0 tasks   \cite{fu2020d4rl}, 
comparing with several baselines that augment data for policy training:  
\begin{itemize}[topsep=0pt, partopsep=0pt, leftmargin=*] 
\item \textbf{CABI}   \cite{lyu_double_2022}, the SOTA policy-decoupled data augmentation algorithm for ORL. we reproduce CABI following their paper   \cite{lyu_double_2022} based on CORL   \cite{tarasov2022corl}.
\item \textbf{COMBO}    \cite{yu2021combo} and \textbf{MOPO}    \cite{yu2020mopo}, two widely studied model-based ORL methods, which are policy-dependent. we re-run COMBO~  \cite{yu2021combo} using code of   \cite{offinerlkit}, and take the reported results of MOPO  directly from the original paper~  \cite{yu2020mopo}. Additionally, We re-run MOPO on expert datasets using OfflineRL-Kit   \cite{offinerlkit}.
\end{itemize}

To ensure the fairness of the comparison, we implement both HIPODE and the downstream ORL algorithms~(TD3BC and CQL) based on CORL   \cite{tarasov2022corl}. For the implementation of HIPODE, we tune the synthetic rate $\eta$, which represents the proportion of synthetic in each offline policy learning batch, as a main hyper-parameter. Most of the tasks have a similar score from a synthetic rate < 0.5. Besides, we choose action selecting rate $\lambda$ from $\{0.2, 1.0\}$, and set the penalty weight $\alpha$ 1.0 as default. 
We present detailed discussion about benchmark tasks and implementation in Appendix \ref{appendix: settings}. 



\begin{table}
  \caption{Normalized average score and standard deviation over 3 seeds of HIPODE based on downstream ORL algorithms~(CQL and TD3BC) and baseline performance. In the table, -m-e, -m-r, -m, -r, -e denote -medium-expert, -medium-replay, -medium, -random, -expert respectively.}
  \label{mujoco results}
  \centering
  \scalebox{0.82}{
  \begin{tabular}{>{\raggedright}m{2.4cm}|>{\raggedleft}m{1.8cm}>{\raggedleft}m{1.1cm}>{\raggedleft}m{0.9cm}|>{\raggedleft}m{1.8cm}>{\raggedleft}m{1.2cm}>{\raggedleft}m{1.0cm}|>{\raggedleft}m{1.1cm}>{\raggedleft}m{1.1cm}}
    \toprule
    \textbf{Task Name} &\textbf{{CQL\\+ HIPODE}} & \textbf{{CQL\\+ CABI} } & \textbf{CQL}  & \textbf{{TD3BC\\+ HIPODE}} & \textbf{{TD3BC\\+ CABI}} & \textbf{TD3BC} & \textbf{COMBO}   & \textbf{MOPO}  \tabularnewline
    \midrule
   
    halfcheetah-m-e  & $99.5\pm{4.5}$   & $101.0$  & 94.8     & $102.6\pm{2.2}$  & 94.6  & 98.0   &38.7   & 63.3 \tabularnewline
    hopper-m-e       & $112.1\pm{0.1}$  & 112.0 & 111.9        & $112.4\pm{0.3}$  & 102.8 & 112.0  & 75.1  & 23.7\tabularnewline
    walker2d-m-e     & $94.0\pm{4.0}$   & 92.3  & 70.3         & $105.3\pm{4.0} $ & 99.6  & 105.4  & 2.3   & 44.6\tabularnewline
    halfcheetah-m-r  & $46.0\pm{0.1}$   & 42.8  & 42.5         & $44.0\pm{0.7} $  & 43.4  & 43.3   & 46.9  & 53.1\tabularnewline
    hopper-m-r       & $34.7\pm{2.9}$   & 29.7  & 28.2         & $36.2\pm{0.8} $  & 32.7  & 32.7   & 19.7  & 67.5\tabularnewline
    walker2d-m-r     & $21.7\pm{6.5}$   & 12.7  & 5.1          & $36.4\pm{11.1}$   & 37.7  & 19.6  & 19.5  & 39.0\tabularnewline
    halfcheetah-m    & $39.3\pm{0.2}$   & 36.7  & 39.2         & $43.7\pm{0.5}$   & 43.0  & 43.7   & 27.4  & 42.3\tabularnewline
    hopper-m         & $30.4\pm{0.9}$   & 30.5  & 30.3         & $99.9\pm{0.3} $  & 99.7  & 99.9   & 71.6  & 28.0\tabularnewline
    walker2d-m       & $75.7\pm{5.1}$   & 46.8  & 66.9         & $80.1\pm{0.7} $  & 80.3  & 79.7   & 71.8  & 17.8\tabularnewline
    halfcheetah-r    & $23.1\pm{1.5}$   & 2.8   & 17.0         & $15.5\pm{0.9} $  & 12.5  & 12.8   & 5.5   & 35.4\tabularnewline
    hopper-r         & $10.4\pm{0.1}$   & 10.2  & 10.4         & $10.9\pm{0.2} $  & 10.9  &10.9    & 7.5   & 11.7\tabularnewline
    walker2d-r       & $10.5\pm{0.6}$   & -0.1  & 1.7          & $6.6\pm{1.0}  $  & 3.0   & 0.37   & 1.6   & 13.6\tabularnewline\midrule
    \textbf{Total}            & \textbf{597.4}   & 517.4 & 518.3        &  \textbf{693.6}  & 660.2 & 658.3  & 387.5 & 440.0\tabularnewline\midrule   
    halfcheetah-e    & $109.1\pm{0.14}$ & 109.1 & 109.8        & $106.5\pm{0.5}$  & 106.3 & 107.0  & 44.2  & 102.1 \tabularnewline
    hopper-e         & $112.2\pm{0.2}$  & 112.3 & 112.0        & $112.3\pm{0.5}$  & 110.8 & 107.3  & 112.3 & 0.7 \tabularnewline
    walker2d-e       & $109.3\pm{2.0}$  & 98.7  & 104.7        & $106.6\pm{5.2}$  & 102.5 & 98.7   & 37.3  & 2.1\tabularnewline
    \midrule
    \textbf{Total}            & \textbf{929.0}  & 837.5 & 844.9          & \textbf{1019.0} & 979.8 & 971.5  & 581.3 & 544.9 \tabularnewline
    \bottomrule
  \end{tabular}}
\end{table}

\textbf{Main results}. Table \ref{mujoco results} shows the results on 15 MuJoCo tasks comparing between the above-mentioned algorithms. HIPODE achieves remarkable improvements over baselines~(TD3BC, CQL), and also significant gain outperforming SOTA data augmentation method~(CABI), which confirm the effectiveness of HIPODE in handling these offline tasks. In this experiment, in order to make a fair comparison, we reproduce the CABI algorithm based on the same downstream policy learning process and the same hyper-parameters to demonstrate HIPODE's superiority. Under the premise of controlling downstream implementation and consistent hyper-parameters, the advantage of HIPODE performance all comes from the data augmentation process. On the other hand, compared to the reported results in   \cite{lyu_double_2022}, our advantage remains consistent, as detailed in Appendix \ref{appendix: additional results}. HIPODE is also effective in the Adroit tasks. Details are presented in Appendix \ref{appendix: additional results}.

Also, HIPODE's predominance over the model-based policy-dependent baseline algorithms~(COMBO, MOPO) demonstrates its strength. Noting that COMBO and MOPO need to access the true terminal function to ensure algorithm performance, whereas HIPODE achieves better performance without the need for such a function, by uniformly setting terminal flag of HIPODE's synthetic data to False. 

To show HIPODE indeed generates high-quality transitions, we visualize the distribution of estimated discounted cumulative rewards of trajectories in the original dataset and synthetic data generated by HIPODE and CABI. Specifically, we train online SAC to converge to obtain the optimal value 
\begin{wrapfigure}{r}{0.38\textwidth}
	\centering
        \includegraphics[width=0.38\textwidth]{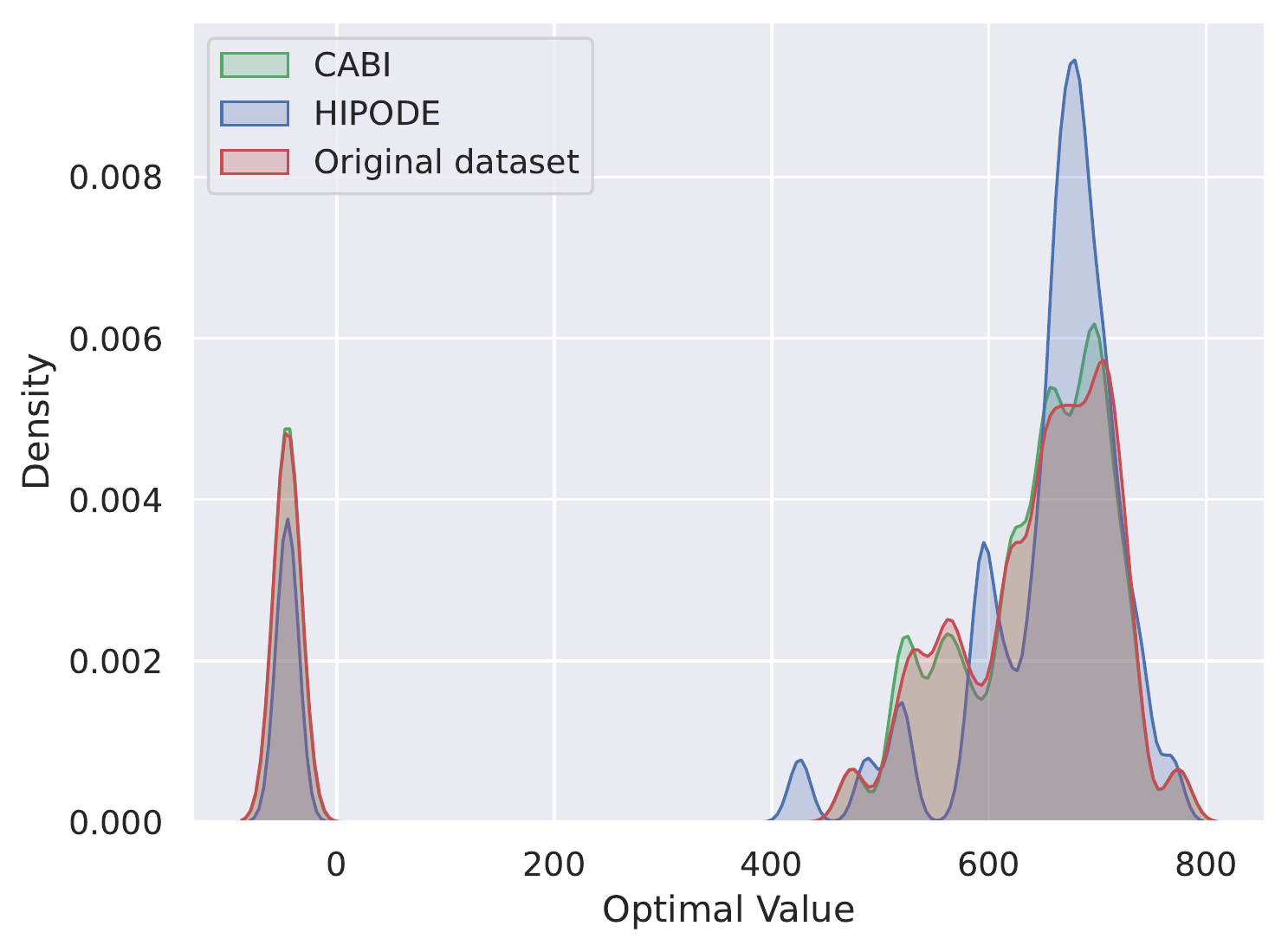}
	\caption{Density of different synthetic data and the original dataset.}
	\label{fig:density}
\end{wrapfigure}
function $V^{*}$ as authoritative value function. For each state-action pair $(s,a)$, we use $r+V^{*}(s')$ to represent the  discounted cumulative reward, where $r,s' \sim p(r,s'|s,a)$ are the true reward and next state in the environment respectively. 
Fig.\ref{fig:density} illustrates the density of synthetic data generated by CABI and HIPODE on halfcheetah-medium-replay-v0, as well as the original dataset, where X axis represents $V^{*}$ and Y represents the density of synthetic transitions on $V^{*}$.
Form the figure, the green shadow almost coincide with the red one, showing that CABI's data distribution almost coincide with the original dataset, while HIPODE indeed generates more high-quality data. In conjunction with the results in Table \ref{mujoco results}, the advantage of HIPIDE performance comes from more high-quality data in augmentation process, which sequentially demonstrates that high-quality data is more suitable rather than high diversity data as augmented data for ORL.

\begin{wraptable}{r}{7.5cm}
\centering
\vspace{-10px}
\caption{Normalized average score of generating different types of augmented data over 3 seeds on MuJoCo -v0 tasks.
}
\vspace{6px}
\label{tab: noV repeat}
\scriptsize
\begin{tabular}{>{\raggedright}m{1.6cm}|>{\raggedleft}m{0.9cm}>{\raggedleft}m{0.9cm}>{\raggedleft}m{0.9cm}>{\raggedleft}m{0.9cm}}

\toprule
\textbf{Task Name}       & \textbf{{Repeat\\+TD3BC}} & \textbf{{NoV\\+TD3BC}}   & \textbf{{HIPODE\\+TD3BC}} & \textbf{TD3BC} \tabularnewline \midrule
halfcheetah-m-e & 97.4         & 99.1        & 102.6        & 98.0  \tabularnewline
hopper-m-e      & 111.9        & 110.5       & 112.4        & 112   \tabularnewline
walker2d-m-e    & 38.0         & 102.5       & 105.3        & 105.4 \tabularnewline
halfcheetah-m-r & 42.5         & 44.0        & 44.0         & 43.3  \tabularnewline
hopper-m-r      & 36.5         & 36.2        & 36.2         & 32.7  \tabularnewline
walker2d-m-r    & 18.0         & 30.0        & 36.4         & 19.6  \tabularnewline
halfcheetah-m   & 43.6         & 43.1        & 43.7         & 43.7  \tabularnewline
hopper-m        & 99.8         & 99.7        & 99.9         & 99.9  \tabularnewline
walker2d-m      & 79.5         & 79.7        & 80.1         & 79.7  \tabularnewline
halfcheetah-r   & 11.7         & 13.1        & 15.5         & 12.8  \tabularnewline
hopper-r        & 11.1         & 10.9        & 10.9         & 10.9  \tabularnewline
walker2d-r      & 1.9          & 2.1         & 6.6          & 0.4   \tabularnewline
halfcheetah-e   & 104.2        & 105.3       & 106.5        & 107.0 \tabularnewline
hopper-e        & 112.5        & 112.3       & 112.3        & 107.3 \tabularnewline
walker2d-e      & 78.1         & 104.9       & 106.6        & 98.7  \tabularnewline \midrule
\textbf{Total}           & 886.8        & 993.3       & \textbf{1019.0}       & 971.4 \tabularnewline \bottomrule
\end{tabular}
\end{wraptable}
\subsection{Ablation Study}\label{ablation study}
In this section, we aim to further investigate how the generated data improves downstream offline policy. We conduct ablation experiments from three perspectives: not generating synthetic data (Repeat), generating vanilla synthetic data (NoV), and generating high-quality synthetic data (HIPODE). and the results are shown in Table \ref{tab: noV repeat}.

Specifically, the difference of Repeat and HIPODE is that the synthetic data is replaced by 10\% high-quality data from the original dataset in Repeat. The difference between NoV and HIPODE is that the value maximization mechanism is removed in NoV, i.e., the quality of generated data is not controlled. Generating high-quality synthetic data is exactly HIPODE. Normalized score of enhancing TD3BC with these perspectives is shown as Repeat+TD3BC, NoV+ TD3BC, HIPODE+TD3BC in Table \ref{tab: noV repeat}.


As the results in Table \ref{tab: noV repeat}, Repeat+TD3BC, i.e., repeating high-quality data in the dataset, brings little performance gain and even hurts performance on walker2d-medium-expert and walker2d-expert. Thus it's not effective for improving ORL performance.
Besides, NoV+TD3BC achieves an improvement over TD3BC, indicating the importance of generating new synthetic data for data augmentation. However, the performance of NoV+TD3BC is worse than HIPODE, indicating the importance of generating high-quality data. To summarize, The result suggests that generating synthetic data is more effective than simply repeat data, but the pursuit of generating higher quality synthetic data can bring more significant performance improvements for downstream ORL performance. 



\begin{table}
\centering
\caption{Normalized score comparison of policy-dependent methods for data augmentation v.s. HIPODE and the baseline on MuJoCo -v0 tasks. We report average normalized score over 3 random seeds each task. Full results consists of -random and -expert tasks are presented in Appendix \ref{appendix: additional results}. }
\label{tab:model based ablation}
\scalebox{0.9}{
\begin{tabular}{@{}l|rrrrr|rr@{}}
\toprule
\textbf{Task Name}            & \textbf{\thead{MB+2.5\\TD3BC}} & \textbf{\thead{MB+0.001\\TD3BC}} & \textbf{MBPO}  & \textbf{\thead{HIPODE+\\TD3BC}} & \textbf{TD3BC} & \textbf{\thead{BooT\\+CQL}} & \textbf{CQL}   \\\midrule
halfcheetah-m-e & 26.0       & 73.0  & 9.7   & 102.6      & 98.0  & 5.0      & 94.8  \\
hopper-m-e      & 1.1        & 42.6  & 56    & 112.4      & 112.0 & 0.8      & 111.9 \\
walker2d-m-e    & 42.9       & 8.5   & 7.6   & 105.3      & 105.4 & 26.4     & 70.3  \\
halfcheetah-m-r & 45.8      & 23.1  & 47.3  & 44         & 43.3  & 4.3      & 42.5  \\
hopper-m-r      & 4.8        & 20.9  & 49.8  & 36.8       & 32.7  & 5.0      & 28.2  \\
walker2d-m-r    & 0.0        & 7.5   & 22.2  & 36.4       & 19.6  & 5.8      & 5.1   \\
halfcheetah-m   & 45.4       & 36.7  & 28.3  & 43.7       & 43.7  & 30.0     & 39.2  \\
hopper-m        & 0.7        & 30.2  & 4.9   & 99.9       & 99.9  & 79.8     & 30.3  \\
walker2d-m      & 4.3        & 16.9  & 12.7  & 80.1       & 79.7  & 6.4      & 66.9  \\\midrule
\textbf{Total}           & 171.0      & 259.4 & 238.5 & 661.2      & 634.3 & 163.5    & 489.2 \\
\bottomrule
\end{tabular}
}
\end{table}

\subsection{Comparison with Policy-Dependent Data Augmentation Methods}
\label{policy-dependent}
In policy-dependent data augmentation methods, the data generation process is tightly tied to the downstream ORL policy, which limits the applicability of the generated data.  In this section we aim to illustrate the strength of our policy-decoupled data augmentation method, compared to policy-dependent methods on different downstream ORL policies. Specifically, on the downstream TD3BC algorithm, we evaluate the effect of data generated with some model-based policy dependent algorithms; on the downstream CQL algorithm, we analyze the effect of the more advanced policy dependent algorithm Boot \cite{wang2022bootstrapped}.

We first evaluate the performance of dynamics-model-enhanced TD3BC based on the data generated by a previously trained dynamics model, by rolling-out current TD3BC policy on the dynamics model. The results are reported as MB+$\alpha$TD3BC in Table \ref{tab:model based ablation}, where $\alpha$ is a hyper-parameter in TD3BC   \cite{fujimoto_minimalist_2021}. 
We also compare offline MBPO with HIPODE since it can also be seen as a method directly using dynamics-model-generated data as augmented data. The difference between model based TD3BC and MBPO is that TD3BC has a behaviour cloning restrict on it's critic    \cite{fujimoto_minimalist_2021} while MBPO    \cite{janner2019trust} dose not. Results in Table \ref{tab:model based ablation} indicate that 
using dynamics-model-generated data as augmentation will damage the offline agent, and such damage can be mitigated when the policy of the offline agent is closed to the behavioural policy of the dataset. This suggests that the damage is caused by the difference between the policy the dynamics model trained on and the policy it generates data on, which these model-based policy-dependent methods fail to address.

We then directly take results report in   \cite{wang2022bootstrapped} to form the BooT+CQL column in Table \ref{tab:model based ablation}.  BooT+CQL means directly using synthetic data generated by BooT on CQL. The results show that synthetic data generated by Boot has poor results as augmented data combined with CQL. This indicates that synthetic data generated by a policy-dependent data augmentation method can damage another offline agent. In contrast, HIPODE is policy-decoupled and our augmented data can benefit different offline agent without changing, as shown in Table \ref{mujoco results}. 

In summary, synthetic data generated by policy-dependent data augmentation methods may have a detrimental effect on ORL processes, while the synthetic data generated by HIPODE can improve their performance, demonstrating the superiority of HIPODE.

\section{Conclusions and Limitations}
In this paper, we investigate the issues of data augmentation for ORL. We conduct extensive experiments to demonstrate that, in the context of ORL, high-quality data is a more suitable choice for augmented data than high-diversity data when the authority of the data is the same. Based on this observation, we propose a novel data augmentation method called HIPODE, which selects states with higher values as augmented data. This ensures that the synthetic data is both authentic and of high-quality and is generated in a policy-decoupled manner. Our experimental results on D4RL benchmarks demonstrate that HIPODE significantly improves the performance of several widely used model-free ORL baselines without changing the augmented data, thereby achieving policy-decoupled data augmentation and demonstrating superiority over policy-dependent methods. Furthermore, HIPODE outperforms SOTA policy-decoupled data augmentation methods for ORL, demonstrating the benefits by generating high-quality data.

The limitation of our work lies in the complexity of the method, as it requires several models to generate synthetic data. Additionally, HIPODE is outperformed by vanilla model-based ORL methods (e.g., MBPO) on -random datasets because the value penalty is excessively strict on those datasets. We believe that adjusting the penalty weight to be state-dependent instead of initially setting it to a fixed value is a potential solution to this issue, which we leave for future work.   

\bibliographystyle{plain}
\bibliography{cite}

\clearpage
\appendix

\section{Detailed Settings of Experiments}\label{appendix: settings}
\subsection{D4RL Tasks}
In this section, we describe details about 
 the MuJoCo and Adroit tasks in the D4RL  \cite{fu2020d4rl} benchmark suite, on which we evaluate HIPODE.

\textbf{MuJoCo} contains a series of continuous locomotion tasks. Among them, Walker2d is a bipedal robot control task, where the goal is to maintain the balance of robot body and move as fast as possible. Hopper is a single-legged robot control task where the goal is to make the robot jump as fast as possible. Halfcheetah is to make a simulated robot perform a running motion that resembles a cheetah's movement, while trying to maximize the distance traveled within a fixed time period. In the MuJoCo-v0 tasks of D4RL, each environment has 5 different types of datasets: expert, medium-expert, medium-replay, medium and random. \textbf{Expert}: a large amount of data collected by a well-trained SAC agent. \textbf{Medium}: a large amount of data collected by a early-stopped SAC agent. \textbf{Medium-expert}: a large amount of mixed data of medium and expert at a 50-50 ratio. \textbf{Medium-replay}: replay buffer of a early-stopped SAC agent. \textbf{Random}: a large amount of data collected by a random policy.

\textbf{Adroit} contains a series of continuous and sparse-reward robotic environment to control a 24-DoF simulated Shadow Hand robot to twirl a pen, hammer a nail, open a door or grab a ball. These environments are even hard for online learning due to sparse rewards and exploration challenges \cite{fu2020d4rl}. There are three types of datasets for each environment: expert, human and cloned. Among them, we evaluate HIPODE on human datasets, where a small number of demonstrations operated by a human (25 trajectories per task) is collected.  

We report normalized score based on the protocol described in \cite{fu2020d4rl}. A score of 0 represents the average return of random policies, and a score of 100 represents the return of a domain-specific expert. In our experiments, all score is the final performance of the downstream offline reinforcement learning (ORL) algorithms, which is the average cumulative reward of ten final policy rollouts. 
\begin{table}[ht]
  \caption{Hyper-parameters of HIPODE+CQL and HIPODE+TD3BC.}
  \label{CQL v0 hypers}
  \centering
  \begin{tabular}{l|rrrrr|r}
    \toprule
    \textbf{\thead{HIPODE+CQL\\Task Name}}       & \textbf{\thead{synthetic\\ rate $\eta$}}  & \textbf{\thead{selecting\\ rate $\lambda$}}    & \textbf{\thead{candidate\\ number $n$}}    & \textbf{\thead{penalty \\weight $\alpha$}}  & \textbf{\thead{penalty\\ scope $\sigma$}} & \textbf{\thead{CQL\\ $\alpha$}}     \\
    \midrule
    halfcheetah-e        & 0.2 & 0.2      & 10     & 1.0 & 1.0 & 10.0\\
    halfcheetah-m-e  & 0.1 & 0.2      & 10     & 1.0 & 1.0 & 10.0\\
    halfcheetah-m-r  & 0.2  & 1.0     & 10     & 1.0 & 1.0 & 10.0\\
    halfcheetah-m         & 0.2  & 1.0 & 10     & 1.0 & 1.0 & 10.0\\
    halfcheetah-r         & 0.1  & 0.2 & 10     & 1.0 & 1.0 & 10.0\\
    hopper-e              & 0.2 & 1.0 & 10     & 1.0 & 1.0 & 10.0\\
    hopper-m-e       & 0.2 & 0.2 & 10     & 1.0 & 1.0 & 10.0\\
    hopper-m-r       & 0.2  & 1.0 & 10     & 1.0 & 1.0 & 10.0\\
    hopper-m              & 0.2  & 0.2 & 10     & 1.0 & 1.0 & 10.0\\
    hopper-r              & 0.2  & 1.0 & 10     & 1.0 & 1.0 & 10.0\\
    walker2d-e            & 0.2 & 0.2 & 10     & 1.0 & 1.0 & 10.0 \\
    walker2d-m-e     & 0.1 & 1.0 & 10     & 1.0 & 1.0 & 10.0\\
    walker2d-m-r     & 0.2  & 1.0 & 10     & 1.0 & 1.0 & 10.0\\
    walker2d-m            & 0.2  & 1.0 & 10     & 1.0 & 1.0 & 10.0\\
    walker2d-r            & 0.2  & 1.0 & 10     & 1.0 & 1.0 & 10.0\\ 
    \bottomrule
  \end{tabular}
  \vskip -0.2in
\end{table}
\subsection{Hyper-Parameters}\label{appendix: hyper}
In this section, we provide the key hyper-parameters used for HIPODE in the main experiments. They are listed in Table \ref{CQL v0 hypers} and Table \ref{td3bc v0 hypers}. Unless specified otherwise, all results in this paper are obtained using the hyper-parameters listed here. To reproduce CABI, we follow the hyper-parameters provided in \cite{lyu_double_2022}. For COMBO, we use the Offline-RL-Kit's code \cite{offinerlkit} and follow \cite{yu2021combo} to set the real ratio to 0.8 for both walker2d-medium-expert and walker2d-expert while 0.5 for other tasks. For all other hyper-parameters in COMBO and MOPO -expert, we use the default hyper-parameters provided in the Offline-RL-Kit's code \footnote{CORL code URL: \url{https://github.com/tinkoff-ai/CORL}}.

\begin{table}
  \caption{Hyper-parameters of HIPODE+TD3BC on 15 MuJoCo -v0 tasks and 4 Adroit-human-v0 tasks.}
  \label{td3bc v0 hypers}
  \centering
  \begin{tabular}{l|rrrrr|r}
    \toprule
    \textbf{Task Name}       & \textbf{\thead{synthetic\\ rate $\eta$}}  & \textbf{\thead{selecting\\ rate $\lambda$}}    & \textbf{\thead{candidate\\ number $n$}}    & \textbf{\thead{penalty \\weight $\alpha$}}  & \textbf{\thead{penalty\\ scope $\sigma$}} & \textbf{\thead{TD3BC\\ $\alpha$}}    \\
    \midrule
    halfcheetah-e         & 0.15 & 0.2 & 10     & 1.0 & 1.0 & 2.5\\
    halfcheetah-m-e  & 0.15 & 0.2 & 10     & 1.0 & 1.0 & 2.5\\
    halfcheetah-m-r  & 0.5  & 0.2 & 10     & 1.0 & 1.0 & 2.5\\
    halfcheetah-m         & 0.2  & 0.2 & 10     & 1.0 & 1.0 & 2.5\\
    halfcheetah-r         & 0.7  & 0.2 & 10     & 1.0 & 1.0 & 2.5\\
    hopper-e              & 0.2 & 0.2 & 10     & 1.0 & 1.0 & 2.5\\
    hopper-m-e       & 0.5 & 0.2 & 10     & 1.0 & 1.0 & 2.5\\
    hopper-m-r       & 0.5  & 0.2 & 10     & 1.0 & 1.0 & 2.5\\
    hopper-m              & 0.1  & 0.2 & 10     & 1.0 & 1.0 & 2.5\\
    hopper-r              & 0.15  & 0.2 & 10     & 1.0\ & 1.0 & 2.5\\
    walker2d-e          & 0.2 & 0.2 & 10     & 1.0 & 1.0 & 2.5\\
    walker2d-m-e     & 0.05 & 0.2 & 10     & 1.0 & 1.0 & 2.5\\
    walker2d-m-r     & 0.5  & 0.2 & 10     & 1.0 & 1.0 & 2.5\\
    walker2d-m            & 0.15  & 0.2 & 10     & 1.0 & 1.0 & 2.5\\
    walker2d-r            & 0.4  & 0.2 & 10     & 1.0  & 1.0 & 2.5\\
    \midrule
    door-human      & 0.4 & 0.2 & 10     & 1.0 & 1.0 & 0.001\\
    hammer-human    & 0.2 & 0.2 & 10     & 1.0 & 1.0 & 0.001\\
    pen-human       & 0.8 & 0.2 & 10     & 1.0 & 1.0 & 0.001\\
    relocate-human  & 0.4 & 0.2 & 10     & 1.0 & 1.0 & 0.001\\
    \bottomrule
  \end{tabular}
\end{table}

\subsection{Implementation Details}
In this section, we describe details of the implementation to our experiments. 

\textbf{Data compression.} For data augmentation, generating and storing large amount of synthetic data into a static dataset before the downstream ORL process can be resource-intensive. To address this issue, in practice, HIPODE is integrated into the downstream ORL process by generating synthetic data in real-time during the ORL process. This technique compress the size of synthetic data to the parameters of several generative models in HIPODE. In the data generating process, the synthetic rate $\eta$ denotes the rate of synthetic data in every batch fed into downstream ORL algorithms. Therefor, a batch of size $N$ contains $\eta N$ synthetic transitions and $(1-\eta)N$ real transitions. 

\textbf{Models in HIPODE.} Here we describe the details about the models in HIPODE. There are 5 independent models in HIPODE:
\begin{itemize}
    \item \textbf{Value network.} The value network is implemented using a Multi-Layer Perceptron (MLP) with one hidden layer of 256 units. We update the target value network every 2 gradient steps using the soft update method $\bar{\theta}=\tau\theta+(1-\tau)\bar{\theta}$, where $\tau=0.005$ is the update rate.
    \item \textbf{Inverse action model.} The inverse action model is implemented using a CVAE \cite{sohn2015learning} to generate an action from a given state and next state. The encoder and decoder of the CVAE both have one hidden layer of 750 units, and the latent dimension is twice the state dimension of each task.
    \item \textbf{Inverse reward model.} The inverse reward model is implemented using a CVAE to generate a reward from a given state and next state. Its structure is similar to the inverse action model, but the two models are trained separately. Together, they are referred to as the `inverse dynamics model'.
    \item \textbf{Forward dynamics model.} The forward dynamics model predicts the next state from a given current state and action. For a fair comparison, the implementation of the forward dynamics model is identical for both CABI and HIPODE and we refer to the implementation of the dynamics model in the D3RLPY \cite{d3rlpy} library \footnote{D3RLPY code URL: \url{https://github.com/takuseno/d3rlpy}} for the implementation of this part.
    \item \textbf{State transition model.} The state transition model is implemented using a CVAE to generate the next state from a given current state, following \cite{zhang2022state}. Its structure is the same as that of the inverse action model, except that the input dimension of the encoder and the output dimension of the decoder are different. 
\end{itemize}
From the details of the HIPODE models, it can be seen that, although the data generating process is integrated into the downstream ORL process, the training process and data generating process of all the models in HIPODE is decoupled from the downstream ORL process, thus achieving policy-decoupled data augmentation. 

\begin{wraptable}{r}{0.55\textwidth}
\vskip -0.2in
    \centering
\caption{Average normalized score and standard deviation of HIPODE+TD3BC v.s. TD3BC over 3 random seeds on Adroit-human tasks.}
\label{tab:adroit}
\vskip 0.1in
\begin{tabular}{@{}l|rr@{}}
\toprule
\textbf{Task Name}      & \textbf{HIPODE+TD3BC} & \textbf{TD3BC} \\ \midrule
door-human     & 2.7 $\pm$ 2.6         & 1.3 $\pm$ {1.2}  \\
hammer-human   & 3.5 $\pm$ 4.0          & 3.9 $\pm$ 5.6  \\
pen-human      & 85.0 $\pm$ 14.3        & 60.8 $\pm$ 14.0\\
relocate-human & 0.2 $\pm$ 0.1        & 0.1 $\pm$ 0.1  \\ \midrule
\textbf{Total}          & \textbf{91.4}         & 66.1  \\ \bottomrule
\end{tabular}
\vskip -0.1in
\end{wraptable}

\section{Additional Results}\label{appendix: additional results}
\subsection{HIPODE on Adroit Tasks}
To further demonstrate the effectiveness of HIPODE, we also conduct experiments on Adroit-human tasks to evaluate its performance in a more challenging setting with limited human demonstrations and sparse rewards.

 As shown in Table \ref{tab:adroit}, HIPODE is effective, improving the baseline TD3BC on three out of four challenging Adroit tasks. However, the performance of HIPODE+TD3BC on Adroit-human datasets is less effective. We attribute this to the poor performance of the baseline offline agent on Adroit tasks due to the tasks' complexity \cite{fu2020d4rl}.

\subsection{HIPODE V.S. Reported CABI Results}\label{appendix: reported CABI}

In Table \ref{mujoco results}, we report score of reproduced CABI for a fair comparison. However, our reproduced results are slightly worse than those reported in the original CABI paper. To provide a more comprehensive view of performance, we list the results reported in the original CABI paper in Table \ref{tab:reported CABI}. As shown in Table \ref{tab:reported CABI}, HIPODE's advantage compared to CABI is still significant when combined with CQL, while comparable when combined with TD3BC.

Additionally, to further demonstrate the strong potential of HIPODE, we also conduct experiments tuning the penalty weight $\alpha$. We find that adjusting the penalty weight $\alpha$ on some tasks (e.g. walker2d-expert-v0 in Table \ref{tab:penalty weight ablation}) can improve the performance, leading to a higher total score than that of reported CABI+TD3BC. For the sake of fair comparison and ease of use, we only report the results of penalty-weight-non-tuned experiments in this paper. 
\begin{table}[h]
\centering
\caption{Average normalized score of HIPODE over 3 random seeds v.s. reported CABI on 15 MuJoCo -v0 tasks.}
\label{tab:reported CABI}
\begin{tabular}{@{}l|rr|rr@{}}
\toprule
\textbf{Task Name}       & \textbf{\thead{HIPODE\\+CQL}} & \textbf{\thead{CABI\\+CQL}} & \textbf{\thead{HIPODE\\+TD3BC}} & \textbf{\thead{CABI\\+TD3BC}} \\ \midrule
halfcheetah-m-e & 99.5       & 35.3     & 102.6        & 105.0      \\
hopper-m-e      & 112.1      & 112.0    & 112.4        & 112.7      \\
walker2d-m-e    & 94.0       & 107.5    & 105.3        & 108.4      \\
halfcheetah-m-r & 46.0       & 44.6     & 44.0         & 44.4       \\
hopper-m-r      & 34.7       & 34.8     & 36.2         & 31.3       \\
walker2d-m-r    & 21.7       & 21.4     & 36.4         & 29.4       \\
halfcheetah-m   & 39.3       & 42.4     & 43.7         & 45.1       \\
hopper-m        & 30.4       & 57.3     & 99.9         & 100.4      \\
walker2d-m      & 75.7       & 62.7     & 80.1         & 82.0       \\
halfcheetah-r   & 23.1       & 30.2     & 15.5         & 15.1       \\
hopper-r        & 10.4       & 10.7     & 10.9         & 11.9       \\
walker2d-r      & 10.5       & 7.3      & 6.6          & 6.4        \\
halfcheetah-e   & 109.1      & 99.2     & 106.5        & 107.6      \\
hopper-e        & 112.2      & 112.0    & 112.3        & 112.4      \\
walker2d-e      & 109.3      & 110.2    & 106.6        & 108.6      \\ \midrule
\textbf{Total}           & \textbf{928.0}      & 887.6    & 1019.0       & 1020.7     \\ \bottomrule
\end{tabular}
\vskip -0.1in
\end{table}

\subsection{Full Results for Policy-Dependent Methods V.S. HIPODE}\label{appendix: mb}
In this section, we present the results of all 15 MuJoCo tasks to compare the performance of HIPODE with policy-dependent data augmentation methods, and the results are shown in Table \ref{tab:full model based ablation}. 
We begin by evaluating the performance of dynamics-model-enhanced TD3BC, using data generated by a previously trained dynamics model. Specifically, we roll out the current TD3BC policy on the dynamics model. The results are reported in Table \ref{tab:full model based ablation} as MB+$\alpha$TD3BC, where $\alpha$ is a hyper-parameter in TD3BC \cite{fujimoto_minimalist_2021}. 
The  MB+$\alpha$TD3BC results, together with those from MBPO, suggest that using dynamics-model-generated data as augmentation can harm the offline agent. However, we find that the damage can be reduced when the policy of the offline agent is close to the behavioural policy of the dataset. This suggests that the damage is caused by the mismatch between the policy that the dynamics model is trained on and the policy it uses to generate data, which cannot be addressed by model-based policy-dependent methods. 

We then conduct experiments on a more advanced policy-dependent data augmentation method, BooT \cite{wang2022bootstrapped}. We directly take results report in   \cite{wang2022bootstrapped} to form the BooT+CQL column in Table \ref{tab:model based ablation}. BooT+CQL presents the use of synthetic data generated by BooT as augmentation data for CQL. The results indicate that the synthetic data generated by BooT performs poorly when used as augmentation data combined with CQL.This suggests that synthetic data generated by a policy-dependent data augmentation method can have a detrimental effect on ORL algorithms. In contrast, HIPODE's synthetic data can benefit them without causing harm, as demonstrated in the HIPODE+TD3BC and HIPODE+CQL column in Table \ref{tab:full model based ablation}. 

In summary, the synthetic data generated by policy-dependent data augmentation methods can have a negative impact on ORL processes, while HIPODE' synthetic data can improve them, demonstrating the superiority of HIPODE.

\begin{table}
\centering
\caption{Full results of normalized score comparison of policy-dependent methods for data augmentation v.s. HIPODE and the baseline. We report average score over 3 random seeds each task.}
\label{tab:full model based ablation}
\scalebox{0.9}{
\begin{tabular}{@{}l|lllll|rrc@{}}
\toprule
Task Name            & \textbf{\thead{mb+2.5\\TD3BC}} & \textbf{\thead{mb+0.001\\TD3BC}} & \textbf{MBPO}  & \textbf{\thead{HIPODE+\\TD3BC}} & \textbf{TD3BC} & \textbf{\thead{BooT\\+CQL}} & \textbf{CQL}  & \textbf{\thead{HIPODE+\\CQL}} \\\midrule
halfcheetah-m-e & 26.0       & 73.0  & 9.7   & 102.6      & 98.0  & 5.0      & 94.8  &99.5\\
hopper-m-e      & 1.1        & 42.6  & 56    & 112.4      & 112.0 & 0.8      & 111.9 &112.1\\
walker2d-m-e    & 42.9       & 8.5   & 7.6   & 105.3      & 105.4 & 26.4     & 70.3  &94.0\\
halfcheetah-m-r & 45.75      & 23.1  & 47.3  & 44         & 43.3  & 4.3      & 42.5  &46.0\\
hopper-m-r      & 4.8        & 20.9  & 49.8  & 36.8       & 32.7  & 5.0      & 28.2  &34.7\\
walker2d-m-r    & 0.0        & 7.5   & 22.2  & 36.4       & 19.6  & 5.8      & 5.1   &21.7\\
halfcheetah-m   & 45.4       & 36.7  & 28.3  & 43.7       & 43.7  & 30.0     & 39.2  &39.3\\
hopper-m        & 0.7        & 30.2  & 4.9   & 99.9       & 99.9  & 79.8     & 30.3  &30.4\\
walker2d-m      & 4.3        & 16.9  & 12.7  & 80.1       & 79.7  & 6.4      & 66.9  &75.7\\\midrule
Total           & 171.0      & 259.4 & 238.5 & \textbf{661.2}      & 634.3 & 163.5    & 489.2 &553.4\\\midrule
halfcheetah-r   & 27.2       & 2.3   & 30.7  & 15.5       & 12.8  & -        & 17.0  &23.1\\
hopper-r        & 4.7        & 9.6   & 4.5   & 10.9       & 10.9  & -        & 10.4  &10.4\\
walker2d-r      & 0.1        & 1.3   & 13.6  & 6.4        & 0.4   & -        & 1.7   &10.5\\\midrule
Total           & 203.0      & 272.6 & 287.3 & \textbf{694.0}      & 658.4 &          & 518.3 &597.4\\\midrule
halfcheetah-e   & -2.1       & 106.8 & -     & 106.5      & 107.0 & -        & 109.8 &109.1\\
hopper-e        & 1.3        & 111.9 & -     & 112.3      & 107.3 & -        & 112.0 &112.2\\
walker2d-e      & 49.3       & 84.4  & -     & 106.6      & 98.7  & -        & 104.7 &109.3\\\midrule
Total           & 251.0      & 575.7 & -     & \textbf{1019.4}     & 971.4 & -        & 844.9 &929.0 \\
\bottomrule
\end{tabular}
}
\end{table}


\subsection{More Ablation Study}\label{appendix:ablation study}
In this section, we investigate how HIPODE enhances downstream offline policy learning performance by examining two key components: the negative sampling mechanism and the state transition model. We analyze the impact of these components as independent variables of downstream offline policy learning performance.

\textbf{Is the state transition model critical?} To investigate the importance of state transition model, we remove the state transition model and randomly generate candidate next states inside the hypercube formed by the states in the original dataset with the other mechanisms and hyper-parameters stay the same. Results in Table \ref{tab:no stm ablation} shows that removing state transition model severely drops compare to the baseline. 
In terms of the results in Table \ref{tab:no stm ablation}, although the negative sampling mechanism penalizes the OOD states in the no-state-transition-model condition, randomly choosing candidate next states can damage the downstream offline policy learning process, demonstrating the necessity of a state transition model. We believe the following reasons are responsible for this: 
\begin{wraptable}{r}{7.5cm}
\caption{Normalized results over 3 random seeds of HIPODE with randomly sampling next state v.s. HIPODE with CVAE next state.}
\label{tab:no stm ablation}
\vskip 0.1in
\centering
\scalebox{0.8}{
\begin{tabular}{llll}
\toprule           
Name            & \textbf{\thead{HIPODE(no stm)\\+TD3BC}} & \textbf{\thead{HIPODE+\\TD3BC}} & TD3BC \\\midrule
halfcheetah-m-r & 0.6$\pm$0.1                & 44         & 43.3  \\
halfcheetah-m   & 27.3$\pm$2.9               & 43.7       & 43.7  \\
halfcheetah-m-e & 58.2$\pm$3.0               & 102.6      & 98.0  \\
\bottomrule
\end{tabular}
}
\end{wraptable}
(1) The value function is not authentic on randomly sampled next states, so the value is not very effective; (2) few in-support next states are generate so the winning next state may still be an OOD state resulting in a inauthentic synthetic transition; (3) The synthetic transition can lead the policy to a randomly state during evaluation. Hence, state transition model is significantly critical to ensure an authentic augmentation. This resembles the necessity of a behavioral policy in CABI and is consistent with their conclusions    \cite{lyu_double_2022}. 

\textbf{Is negative sampling critical?}
In negative sampling, the penalty weight controls the severity of the penalty added to the value of OOD states. A larger penalty weight makes it less likely for HIPODE to generate an OOD next state. We change the penalty weight in $\{-1, 0, 1, 2, 4, 8\}$ and run the downstream offline policy learning algorithm, without changing the other hyper-parameters on walker2d-expert-v0 and halfcheetah-medium-replay-v0. The results in Table \ref{tab:penalty weight ablation} show that penalty weights greater than 0 outperforms the others, but overall, the score difference is marginal. This indicates that penalizing the value function on OOD states can indeed benefit downstream offline policy learning process. On the other hand, the state transition model generates candidate states near the dataset, which makes the effect of the penalty insignificant.

\begin{table}[]
\centering
\caption{Normalized score of TD3BC combined with different penalty weight for data augmentation. The numbers in the headline denotes the penalty weight. We report average score over 3 random seeds each task.}
\label{tab:penalty weight ablation}
\label{tab:my-table}
\scalebox{0.8}{
\begin{tabular}{@{}l|ccccccc@{}}
\toprule
\textbf{Task Name} &\textbf{-1+\thead{TD3BC}} &\textbf{0+\thead{TD3BC}} &\textbf{1+\thead{TD3BC}} &\textbf{2+\thead{TD3BC}} &\textbf{4+\thead{TD3BC}} &\textbf{8+\thead{TD3BC}} &\textbf{\thead{TD3BC}} \\ \midrule
walker2d-e          &105.2$\pm$2.1 &102.9$\pm$8.1  &106.6$\pm$5.2   &105.2$\pm$2.2  &106.6$\pm$1.0  &109.1$\pm$0.2 & 98.7           \\
halfcheetah-m-r     &43.0$\pm$0.3  &40.6$\pm$2.2   &44.0$\pm$0.4    &44.0$\pm$ 0.7  &43.8$\pm$0.6   &43.8$\pm$0.3 & 43.3          \\ \bottomrule
\end{tabular}
}
\end{table}

\end{document}